\documentclass[conference]{IEEEtran}
\ifCLASSINFOpdf
\else
\fi
\usepackage{graphicx}
\usepackage{tikz}
\usepackage{amsmath}
\usepackage{xcolor}
\usepackage{booktabs}
\usepackage{multirow}
\DeclareMathOperator*{\otherwise}{otherwise}
\DeclareMathOperator*{\onsdp}{on \ SDP}

\begin{document}
%
\title{A Global-Local Attention Mechanism for Relation Classification}

\author{\IEEEauthorblockN{Yiping SUN*}
\IEEEauthorblockA{School of Electronic Information\\ and Electrical Engineering\\ Shanghai JiaoTong University\\ Shanghai, China
}
}


%


\maketitle

\begin{abstract}
Relation classification, a crucial component of relation extraction, involves identifying connections between two entities. Previous studies have predominantly focused on integrating the attention mechanism into relation classification at a global scale, overlooking the importance of the local context. To address this gap, this paper introduces a novel global-local attention mechanism for relation classification, which enhances global attention with a localized focus. Additionally, we propose innovative hard and soft localization mechanisms to identify potential keywords for local attention. By incorporating both hard and soft localization strategies, our approach offers a more nuanced and comprehensive understanding of the contextual cues that contribute to effective relation classification. Our experimental results on the SemEval-2010 Task 8 dataset highlight the superior performance of our method compared to previous attention-based approaches in relation classification.
\end{abstract}


%
\IEEEpeerreviewmaketitle

\section{Introduction}
Relation classification, as a subset of relation extraction, has garnered significant interest in recent years due to its utility in discerning semantic relationships between entities. This field finds application in various domains such as information retrieval, question answering, and more. Particularly noteworthy is its role in Retrieval Argument Generation (RAG) for Large Language Models, where the relationships identified form the backbone of what is commonly known as a Knowledge Graph. The utilization of Knowledge Graphs \cite{yang2023chatgpt} in large language models serves to enrich the model's comprehension of information and its contextual associations. These graph-like structures represent relationships between entities as nodes and edges, thereby furnishing the model with a nuanced contextual framework. In processing vast textual datasets and knowledge bases, Knowledge Graphs aid the model in deducing connections between entities, thereby facilitating a deeper and more holistic understanding of the context. This structured representation enhances the model's capability to answer queries, generate coherent responses, and navigate intricate information landscapes.
In formal, we can describe relation classification as follows: Given a natural sentence $x_1,x_2,...,x_n$ with two entities $e_1,e_2$, the system needs to classify the semantic relation $y$ between two entities.

For example: "His niece$_{e1}$ moved into this apartment$_{e2}$ last month." Relation between niece and apartment is Entity-Destination(e1,e2).

In recent studies, attention mechanisms have been incorporated into relation classification, as discussed in \cite{peng2020bg}\cite{qin2017designing}\cite{lu2017instance}. The objective is to emphasize key words on a \textbf{global} scale, involving the consideration of all words in a sentence when calculating attention weights. However, for long or complex sentences, the prevalence of numerous noisy words can hinder the accurate identification of key words.

Take, for instance, the sentence: "After arresting a computer hacker who stole \$2 million from the Federal Reserve, Jake assumes his identity to infiltrate a ring$_{e1}$ of his associates$_{e2}$ and discover their next target." Localizing potential key words within \textit{ring of associates}, it becomes evident that \textit{of} is crucial. However, combining these words with others may introduce noise, affecting attention weight accuracy.

In this paper, we introduce a global-local attention mechanism for relation classification. Within the attention layer, global and local attention are integrated to enhance overall attention performance. Global attention follows the conventional mechanism, calculating attention weights across all words. In contrast, local attention focuses solely on a few potential key words. To identify such words, we propose a hard and soft localization mechanism. Hard localization assumes all words on the shortest dependency path are potential key words, while soft localization relaxes this assumption, treating the shortest dependency path as the supervision signal for the localization network to identify more robust potential key words.

Using the SemEval-2010 Task 8 dataset, our results demonstrate that our method surpasses previous attention methods for relation classification. Further analysis reveals that global-local attention effectively focuses on correct key words.

\begin{figure*}[t]
\centering
\includegraphics[width=0.8\textwidth]{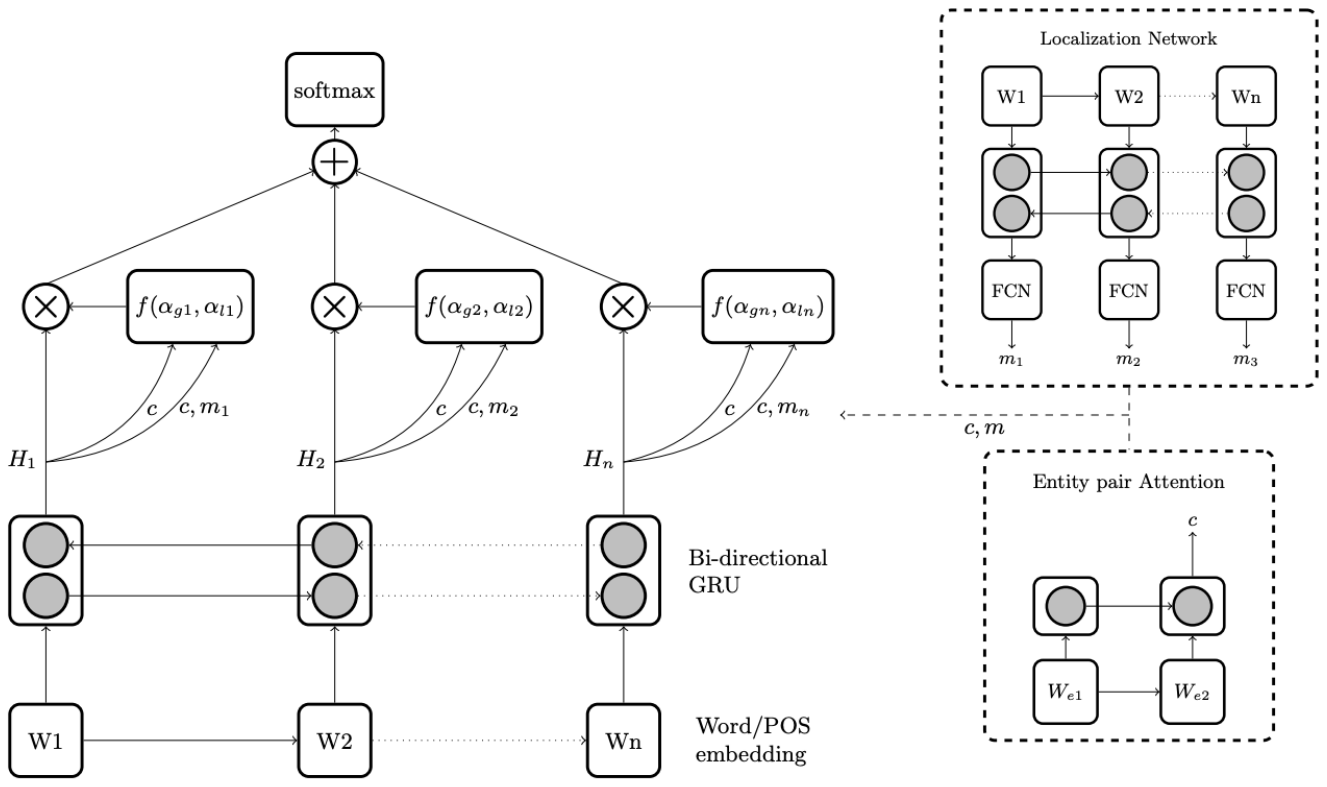}
\caption{Bidirectional GRU network with Global-Local Attention (GLA-BiGRU)}
\label{fig:overview}
\end{figure*}

\section{Related Work}
\subsubsection{Relation Classification}
MVRNN\cite{socher2012semantic} proposed a recursive neural network model that learns compositional vector representations for phrases and sentences of arbitrary syntactic type and length. BiLSTM\cite{zhang2015bidirectional} introduced bidirectional LSTM to let neural network extract features in time dimension. Convolutional neural network method was introduced in CNN\cite{zeng2014relation} and further improved in PCNN\cite{zeng2015distant} by proposing piecewise convolutional neural networks with multi-instance learning. This work\cite{10.1007/978-3-319-99365-2_46} addresses challenges in relation classification by introducing coarse and fine-grained networks that integrate both the entire sentence and key words, enhancing robustness. Additionally, a word selection network under shortest dependency path supervision automatically selects key words, guided by a novel opposite loss for improved feature space alignment.

\subsubsection{Attention in Relation Classification}
Att-BiLSTM\cite{zhou2016attention} first introduced attention mechanism into relation classification field, which reveals topics that better select key words according to attention mechanism. Attention weight of each word is calculated by multiplying the hidden state and attention vector. EAtt-BiGRU\cite{qin2017designing} further improved the attention vector by learning unique attention vector from entity pair information. In this paper, proposed method expand these global attention mechanism into a global-local attention mechanism.


\subsubsection{Global-Local Attention} In Image Caption, \cite{li2017image} proposed a global-local attention (GLA) method by integrating local representation at object-level with global representation at image-level through attention mechanism. In image field, there are various literatures focusing on extracting key subsets of images. But in NLP, it's more difficult for us to localize some potential key words.
In Neural Machine Translation, \cite{luong2015effective} proposed a global approach which always attends to all source words and a local one that only looks at a subset of source words at a time. The local attention is realized by finding a key center in a sentence and expands it by a guassian function. The local attention proposed by \cite{luong2015effective} is unsuitable for relation classification because key words in relation classification may distribute sparsely while in machine translation, target words matches continuous phrases. This minor difference leads to a completely different local attention mechanism and thus a different hybrid method. In this paper, we introduce global-local attention mechanism into relation classification field.

Besides, this article extensively employs several new artificial intelligence technologies, such as \cite{10.1145/3639592.3639600}\cite{wu2024switchtab}\cite{liu2024deep}\cite{zi2024research}\cite{liu2024enhanced}. These technologies have been instrumental in advancing the field of AI and have enabled the development of innovative solutions for various real-world problems. By integrating these cutting-edge techniques, our work aims to address complex challenges and contribute to the ongoing progress in artificial intelligence research and applications.

\section{Methodology}

Fig.\ref{fig:overview} shows the structure of proposed method, which consists of four parts:
\begin{itemize}
\item Input Representation: Words are embedded into word and position embeddings, the latter are based on the relative positions of two entities.
\item Bi-directional GRU Layer: GRU is utilized to extract high-level features from input.
\item Global-Local attention Layer: Attention vector is generated from entity pair information. When generating attention weight, global attention considers all words while local attention only considers few potential key words. Hard and soft localization mechanism provides such words for local attention.
\item Output Layer: Hidden state are summed by attention weight and final classification result is given by softmax.
\end{itemize}

\subsection{Input Representation}
In relation classification task, words are also embedded into vectors like all other NLP tasks do. The input representation is commonly composed of two parts, word embedding and position embedding.
\subsubsection{Word Embedding}
Each word $x_i$ in the input sentence is mapped to a high-dimensional vector $WV^{emb}\in R^{d_e\times|V|}$ where $d_e$ is the dimension of word vector\cite{mikolov2013distributed} and $|V|$ is the vocabulary size of the corpus.
\subsubsection{Position Embedding}
To highlight two entities in the sentence, relation classification further embeds words into position embedding vectors\cite{zeng2014relation}. For each word $x_i$, it will have two relative position indexes, $p_i^1$ and $p_i^2$, of two entities $e_1$ and $e_2$. Then like word embedding, each position index can be mapped to a high-dimensional vector $PV^{emb}\in R^{d_p\times|P|}$ where $d_p$ is the dimension of position vector and $|P|$ is the number of position index. Finally, the overall input representation for each word is 
\begin{equation}
w_i = [(wv_i)^T, (pv_i^1)^T, (pv_i^2)^T]^T
\end{equation}

\subsection{Bi-directional GRU Layer}
We employ Bi-directional GRU\cite{cho2014learning} to extract high-level features. It is a simplified version of LSTM, which only uses update gate $z_t$ and reset gate $r_t$ instead of forget gate, input gate and output gate in LSTM.
\begin{align}
z_t&=\sigma(W_zx_t+U_zh_{t-1})\\
r_t&=\sigma(W_tx_t+U_th_{t-1})\\
\tilde{h_t}&=tanh(Wx_t+U(r_t\odot h_{t-1}))\\
h_t&=(1-z_t)\odot h_{t-1}+z_t\odot \tilde{h_t}
\end{align}
GRU can remain important features in a wide time range and eliminate gradient vanishing problem than LSTM. It is proved to be efficient in relation classification by previous work. The standard GRU is unidirectional. To construct forward and backward information together, we combine forward hidden state $h_i^f\in R^{d_h}$ with backward hidden state $h_i^b\in R^{d_h}$.
\begin{equation}
H_i=[h_i^{f},h_i^{b}]
\end{equation}

\subsection{Bi-directional GRU Layer}
We employ Bi-directional GRU\cite{cho2014learning} to extract high-level features. It is a simplified version of LSTM, which only uses update gate $z_t$ and reset gate $r_t$ instead of forget gate, input gate and output gate in LSTM.
\begin{align}
z_t&=\sigma(W_zx_t+U_zh_{t-1})\\
r_t&=\sigma(W_tx_t+U_th_{t-1})\\
\tilde{h_t}&=tanh(Wx_t+U(r_t\odot h_{t-1}))\\
h_t&=(1-z_t)\odot h_{t-1}+z_t\odot \tilde{h_t}
\end{align}
GRU can remain important features in a wide time range and eliminate gradient vanishing problem than LSTM. It is proved to be efficient in relation classification by previous work. The standard GRU is unidirectional. To construct forward and backward information together, we combine forward hidden state $h_i^f\in R^{d_h}$ with backward hidden state $h_i^b\in R^{d_h}$.
\begin{equation}
H_i=[h_i^{f},h_i^{b}]
\end{equation}

\subsection{Global-Local Attention Mechanism}
\subsubsection{Global Attention}
Entity pair attention is introduced in \cite{qin2017designing}, trying to generating adaptive attention vector using entity pair information. We employ it as our baseline and global attention. Since our method can replace specific attention mechanism with more sophisticated method, ours is a model-free method. 
Let $w_{e1}$ and $w_{e2}$ represent the word/pos embedding of entity1 and entity2. Attention vector $c\in R^{2d_h}$ is the last hidden state generated by a unidirectional GRU using entity pair information. 
\begin{equation}
c = GRU(w_{e1},w_{e2})
\end{equation}

By multiplying each hidden state of BiGRU with attention vector, we can get the global attention weight $\alpha_{gi}$ for each word.
\begin{equation}
\alpha_{gi} = \frac{exp(H_i^Tc)}{\sum_iexp(H_i^Tc)}
\end{equation}

\begin{table*}[t]
\caption {Comparison with other work reported on SemEval 2010 Task 8} \label{tab:compare_other} 
\centering
\resizebox{0.8\textwidth}{!}{%
\begin{tabular}{cll}
\hline
Model & Additional Information & F1 \\ \hline
\begin{tabular}[c]{@{}c@{}}SVM\cite{rink2010utd}\end{tabular}    & \begin{tabular}[c]{@{}l@{}}POS, prefixes, morphological, WordNet, dependency parse,\\ Levin classed, ProBank, FrameNet, NomLex-Plus,\\ Google n-gram, paraphrases, TextRunner\end{tabular} & 82.2                                                \\ \hline
\begin{tabular}[c]{@{}c@{}}SDP-LSTM \cite{xu2015classifying}\end{tabular}       & \begin{tabular}[c]{@{}l@{}}Word/Position embedding\\ +POS+GR+WordNet embeddings\end{tabular}                                                                                                       & \begin{tabular}[c]{@{}l@{}}82.4\\ 83.7\end{tabular} \\ \hline
\begin{tabular}[c]{@{}c@{}}BLSTM\cite{zhang2015bidirectional}\end{tabular}       & \begin{tabular}[c]{@{}l@{}}Word/Position embedding\\ +PF+POS+NER+WNSYN+DEP\end{tabular}                                                                                                            & \begin{tabular}[c]{@{}l@{}}82.7\\ 84.3\end{tabular} \\ \hline
\begin{tabular}[c]{@{}c@{}}SPTree\cite{miwa2016end}\end{tabular}       & \begin{tabular}[c]{@{}l@{}}Word embeddings+DEP\end{tabular}                                                                                                            & \begin{tabular}[c]{@{}l@{}}84.4\end{tabular} \\ \hline
\begin{tabular}[c]{@{}c@{}}Att-BiLSTM\cite{zhou2016attention}\end{tabular}    & \begin{tabular}[c]{@{}l@{}}Word/Position embedding\\ +Position Indicator\end{tabular}                                                                                                                & \begin{tabular}[c]{@{}l@{}}82.7\\ 84.0\end{tabular} \\ \hline
\begin{tabular}[c]{@{}c@{}}EAtt-BiGRU\cite{qin2017designing}\end{tabular}      & Word/Position embedding                                                                                                                                                       & 84.3                                                \\ \hline
\begin{tabular}[c]{@{}c@{}}
\textbf{Proposed method } \\ \textbf{(Hard Localization)}
\end{tabular} & \begin{tabular}[c]{@{}l@{}}Word/Position embedding+DEP \end{tabular} & \begin{tabular}[c]{@{}l@{}}84.7\end{tabular} \\ 
\begin{tabular}[c]{@{}c@{}}
\textbf{Proposed method } \\ \textbf{(Soft Localization)}
\end{tabular} & \begin{tabular}[c]{@{}l@{}}Word/Position embedding+DEP \end{tabular} & \begin{tabular}[c]{@{}l@{}}85.0\end{tabular}\\ \hline
\end{tabular}%
}
\end{table*}

\subsubsection{Local Attention}
When sentence is long and complicated, global attention may be unable to capture correct key words due to noise brought by other unrelated words. If we first localize few potential key words, attention can be enhanced by only considering these words. Let $m_i$ be the localization weight for each word representing whether it's potential key words, then the attention weight is generated by a weighted softmax function:
\begin{equation}
\alpha_{li} = \frac{m_i*exp(H_i^Tc)}{\sum_im_i*exp(H_i^Tc)}
\end{equation}
We propose two mechanism for generating localization weight of each word: hard localization and soft localization.
\begin{itemize}
\item \textbf{Hard Localization}: Although we don't have labeled data representing which words may be key words, shortest dependency path between two entities contains valuable information. We assume that words on shortest dependency path are all potential key words while others are not. Based on it, localization weight $m_i$ can be represented as follows:
\begin{equation}
m_i = sdp_i = 
\left\{  
\begin{array}{ll}  
  1 &  x_i \ \onsdp  \\  
  0 &  \otherwise \\  
\end{array}
\right.
\end{equation}
\item \textbf{Soft Localization}: 
The aforementioned assumption is informal and uncertain in the context of relation classification. Firstly, not all entity pairs can yield a shortest dependency path, and some extracted paths may be inaccurate. Secondly, valuable words essential for classification might lie beyond the confines of the shortest dependency path. To refine the localization mechanism, we replace hard localization with a soft localization facilitated by a localization network. This network takes the original sentence as input and subsequently generates localization weights for each word.
\begin{align}
H_l &= BiGRU_l(x) \\
m &= \sigma(W_{l}H_l+b_{l})
\end{align}
$H_l$ is the hidden state generated by BiGRU in localization network. $W_l$ and $b_l$ are weight and bias of fully connected layer.
Different from hard localization that directly set localization weight using shortest dependency path, soft localization treats it as a supervision signal. Sigmoid loss is equipped to supervise localization network.
\begin{equation}
J_{loc}=-\sum_{j=1}^N\sum_{i=1}^l sdp_i^{(j)}log(m_i^{(j)})+(1-sdp_i^{(j)})log(1-m_i^{(j)})
\end{equation}
$l$ represents the length of sentence. Note that in soft localization, $m_i \in (0,1)$.
\end{itemize}

\subsubsection{Hybrid of Global-Local Attention}
After getting global attention and local attention, we combine them together to enhance whole attention's performance. We define the hybrid function $f$ as follows:
\begin{equation}
\alpha_i = f(\alpha_{gi},\alpha_{li}) = \gamma \alpha_{gi}+(1-\gamma)\alpha_{li}
\end{equation}
$\gamma\in[0,1]$ is hybrid ratio.

\subsection{Output Layer}
Finally, the final sentence representation is the sum of hidden state adjusted by attention weight and classifier is the softmax function.
\begin{equation}
s = \sum\nolimits_i \alpha_iH_i
\end{equation}
\begin{equation}
P(y|x) = softmax(W_cs+b_c)
\end{equation}
\begin{equation}
\hat{y} = argmax_y P(y|x)
\end{equation}
When using soft localization, on the base of cross entropy loss of classification, sigmoid loss of localization network is also added. So the overall loss function is:
\begin{equation}
J = J_{cls}+J_{loc}
\end{equation}

\section{Experiments}
\subsection{Dataset and Experimental Setup}
Experiments are done on SemEval-2010 Task 8 dataset, which has 10717 sentences, including 8000 training examples and 2717 testing examples. It has 9 actual relation classes and an additional class \textit{other}, indicating entity pair not belonging to any relation. We employ official scorer (macro F1) as our evaluation metric, which does not consider class \textit{other} and takes the directionality into consideration.

We use pre-trained word embedding GoogleNews-vectors-negative300 to initialize our word embedding layer. 
The parameters in model are all initialized following Gaussian distribution. The hidden state size $d_h$ of GRU is 100. The hybrid rate $\gamma$ is 0.5. The model is optimized by Adadelta with learning rate 1.0. To overcome the overfitting problem, we apply dropout\cite{srivastava2014dropout} on embedding layer and ultimate layer with rate 0.5. Max-norm $||w||\leq 3$ is also applied every 5 gradient steps. 

\subsection{Experimental Results}
\begin{figure*}[t]
\centering
\includegraphics[width=0.9\linewidth]{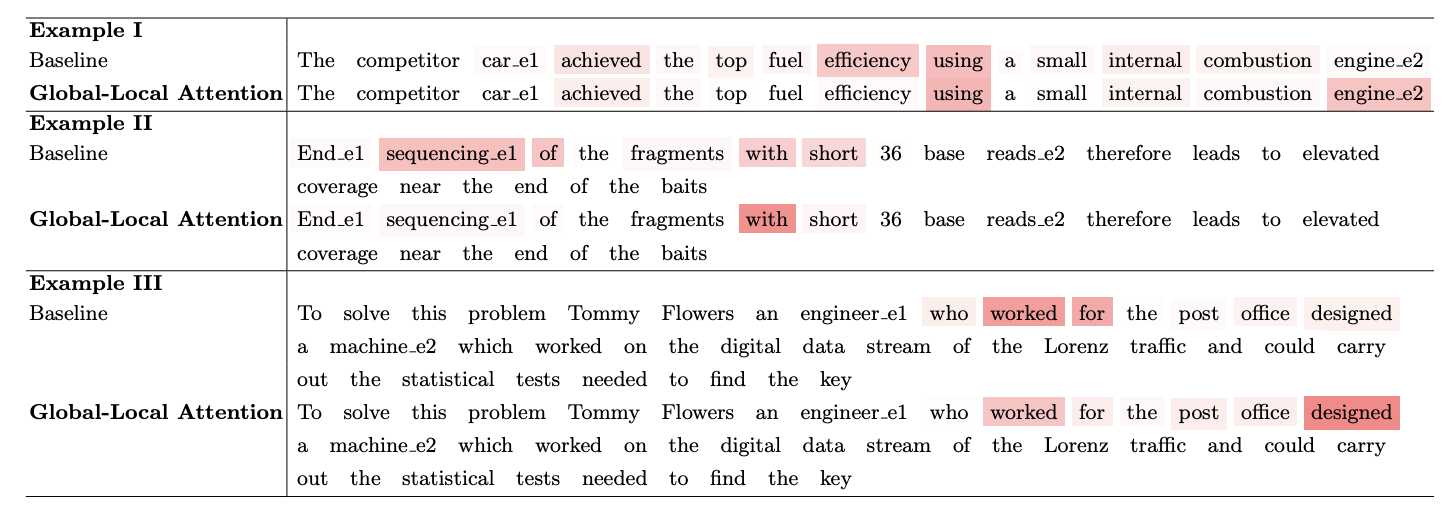}
\caption{COMPARISON OF BASELINE ATTENTION AND GLOBAL-LOCAL
ATTENTION}
\label{fig:analysis}
\end{figure*}
Table.\ref{tab:compare_other} shows comparison between our proposed method and other state-of-the-art systems. 
\begin{itemize}
\item SVM\cite{rink2010utd}: The first top traditional method by equipping a large number of features and support vector machine.
\item SDP-LSTM\cite{xu2015classifying}: Treated shortest dependency path as LSTM's input directly.
\item SPTree\cite{miwa2016end}: Joint extraction of entities and relations under tree-structured LSTM to use dependency parser tree as much as possible.
\item BLSTM\cite{zhang2015bidirectional}: Bidirectional LSTM to extract high-level features.
\item Att-BiLSTM\cite{xu2016improved}: First introduced attention  into relation classification.
\item EAtt-BiGRU\cite{qin2017designing}: Further improved attention mechanism by generating attention weight with prior knowledge of entity pair information.
\end{itemize}

Our baseline is EAtt-BiGRU, so entity pair attention is used to calculate attention vector like they did. But instead of considering all words, proposed method considers words in global and local range together, which leads attention to a better performance. With hard localization, F1 score has been raised to 84.7$\%$. Since hard localization introduce much noise and mistakes on selecting potential key words, with soft localization, F1 score has been improved to the best 85.0$\%$. It proves that soft localization is better than hard localization indirectly. Note that our method doesn't use other lexical features except shortest dependency path.

\subsubsection{Detailed Analysis}
To further evaluate the effectiveness of the proposed global-local attention mechanism, we conducted a comparative analysis with a baseline attention mechanism. Across all three examples, we observed that while the global attention mechanism could correctly identify a few key words, it struggled to distinctly distinguish genuine key words from irrelevant ones. However, with the incorporation of the local attention mechanism, we noticed a significant improvement. Specifically, the attention weights assigned to real key words were increased, while those assigned to noise words were decreased. This enhancement was particularly evident in Example III, where the baseline attention mechanism incorrectly identified \textit{worked for} as key words, whereas the proposed method successfully corrected this error by identifying \textit{designed} as the correct key word. This illustrates the superior performance of our proposed global-local attention mechanism in effectively capturing the most relevant information within the input sequence.

\begin{table}[htbp]
\caption{Comparison of different ratio on combining Global and Local Attention}\label{tab:fusion}
\resizebox{1.0\linewidth}{!}{
\begin{tabular}{|c|c|c|c|c|c|c|c|c|c|c|c}
\hline
$\gamma$&0.0&0.1&0.2&0.3&0.4&\textbf{0.5}\\
\hline
F1-score&84.58&84.59&84.61&84.61&84.72&\textbf{85.04}\\
\hline
\hline
$\gamma$&0.6&0.7&0.8&0.9&1.0&-\\
\hline
F1-score&84.77&84.76&84.70&84.46&84.36&-\\
\hline
\end{tabular}
}
\end{table}

We have also conducted comparisons across different ratios $\gamma$ for combining global and local attention mechanisms. The reported F1-score represents the highest value obtained after conducting $k$ experimental trials. When $\gamma=0.0$, the attention mechanism is purely local, focusing solely on the local context of each word. Conversely, when $\gamma=1.0$, the attention mechanism is purely global, considering the entire input sequence. Interestingly, our findings indicate that local attention slightly outperforms global attention. This observation suggests that while local attention may discard some information and skip updating irrelevant words, global attention is susceptible to noise from irrelevant words. Consequently, our proposed method, which combines both local and global attention, demonstrates greater robustness compared to either local or global attention mechanisms alone.

\section{Conclusion}
In this paper, we propose a global and local attention mechanism for relation classification, which enhances global attention by adding local attention. To localize potential key words, we also propose hard localization and soft localization for local attention.
Result on Semeval 2010 Task 8 shows that under the same feature, our method outperforms previous attention methods for relation classification. Detailed analysis proves that attention becomes more reasonable and robust.
\bibliographystyle{IEEEtran}

%



\end{document}